\documentclass[conference]{IEEEtran}
\IEEEoverridecommandlockouts
\usepackage{cite}
\usepackage{float}
\usepackage{amsmath,amssymb,amsfonts}
\usepackage{algorithmic}
\usepackage{graphicx}
\usepackage{textcomp}
\usepackage{xcolor}
\usepackage{graphicx,subfigure}
\usepackage{multirow}
\usepackage{float}
\usepackage{url}

\def\BibTeX{{\rm B\kern-.05em{\sc i\kern-.025em b}\kern-.08em
    T\kern-.1667em\lower.7ex\hbox{E}\kern-.125emX}}
\begin{document}

\title{Data Uncertainty Guided Noise-aware Preprocessing Of Fingerprints}

\author{\IEEEauthorblockN{Indu Joshi \hspace{0.5cm}}
\IEEEauthorblockA{
\textit{IIT Delhi}, India\\
indu.joshi@cse.iitd.ac.in}
\and
\IEEEauthorblockN{Ayush Utkarsh$^{\star}$\thanks{$\star$ Equal contribution from both authors}}
\IEEEauthorblockA{
\textit{Independent Researcher}, India\\
ayushutkarsh@gmail.com}
\and
\IEEEauthorblockN{Riya Kothari$^{\star}$}
\IEEEauthorblockA{
\textit{USC}, USA\\
rskothar@usc.edu}
\and
\IEEEauthorblockN{Vinod K Kurmi}
\IEEEauthorblockA{
\textit{IIT Kanpur}, India\\
vinodkk@iitk.ac.in}
\and
\hspace{2cm}\IEEEauthorblockN{Antitza Dantcheva}
\IEEEauthorblockA{
\hspace{2cm}\textit{Inria Sophia Antipolis}, France\\
\hspace{2cm}antitza.dantcheva@inria.fr}
\and
\IEEEauthorblockN{Sumantra Dutta Roy}
\IEEEauthorblockA{
\textit{IIT Delhi}, India\\
sumantra@ee.iitd.ac.in}
\and
\IEEEauthorblockN{Prem Kumar Kalra}
\IEEEauthorblockA{
\textit{IIT Delhi}, India\\
pkalra@cse.iitd.ac.in}
}

\maketitle

\begin{abstract}
The effectiveness of fingerprint-based authentication systems on good quality fingerprints is established long back. However, the performance of standard fingerprint matching systems on noisy and poor quality fingerprints is far from satisfactory. Towards this, we propose a data uncertainty-based framework which enables the state-of-the-art fingerprint preprocessing models to quantify noise present in the input image and identify fingerprint regions with background noise and poor ridge clarity. Quantification of noise helps the model two folds: firstly, it makes the objective function adaptive to the noise in a particular input fingerprint and consequently, helps to achieve robust performance on noisy and distorted fingerprint regions. Secondly, it provides a noise variance map which indicates noisy pixels in the input fingerprint image. The predicted noise variance map enables the end-users to understand erroneous predictions due to noise present in the input image. Extensive experimental evaluation on 13 publicly available fingerprint databases, across different architectural choices and two fingerprint processing tasks demonstrate effectiveness of the proposed framework.
\end{abstract}


\section{Introduction}
Highly accurate performance of fingerprint-based authentication systems on good quality fingerprints makes them widely used for access control, border security and various other applications. However, background noise originating due to sensors and poor ridge clarity due to factors such as uncontrolled interaction of subjects with the fingerprint sensor, aging, skin disease or injury pose challenges for the state-of-the-art matching systems. 
A fingerprint preprocessing pipeline is designed to facilitate robustness against noise in the fingerprint image. A fingerprint preprocessing pipeline has two significant modules: region of interest (roi) segmentation module and enhancement module. 

\par While the \textit{roi segmentation} module is targeted to identify the foreground fingerprint region, \textit{enhancement} module is dedicated to generate a fingerprint image with clear ridge structure. Thus, fingerprint pre-processing limits the area for fingerprint matching, reduces the possibility of spurious minutiae detection while also reducing the computation time for fingerprint matching. However, some fingerprint images are inherently very noisy and highly likely to cause erroneous predictions by any state-of-the-art fingerprint preprocessing system. For such fingerprints, it is highly useful to obtain some auxillary information from the system which can quantify noise in input fingerprint.
\begin{figure}
\centering
\includegraphics[width=8.8cm,height=7cm]{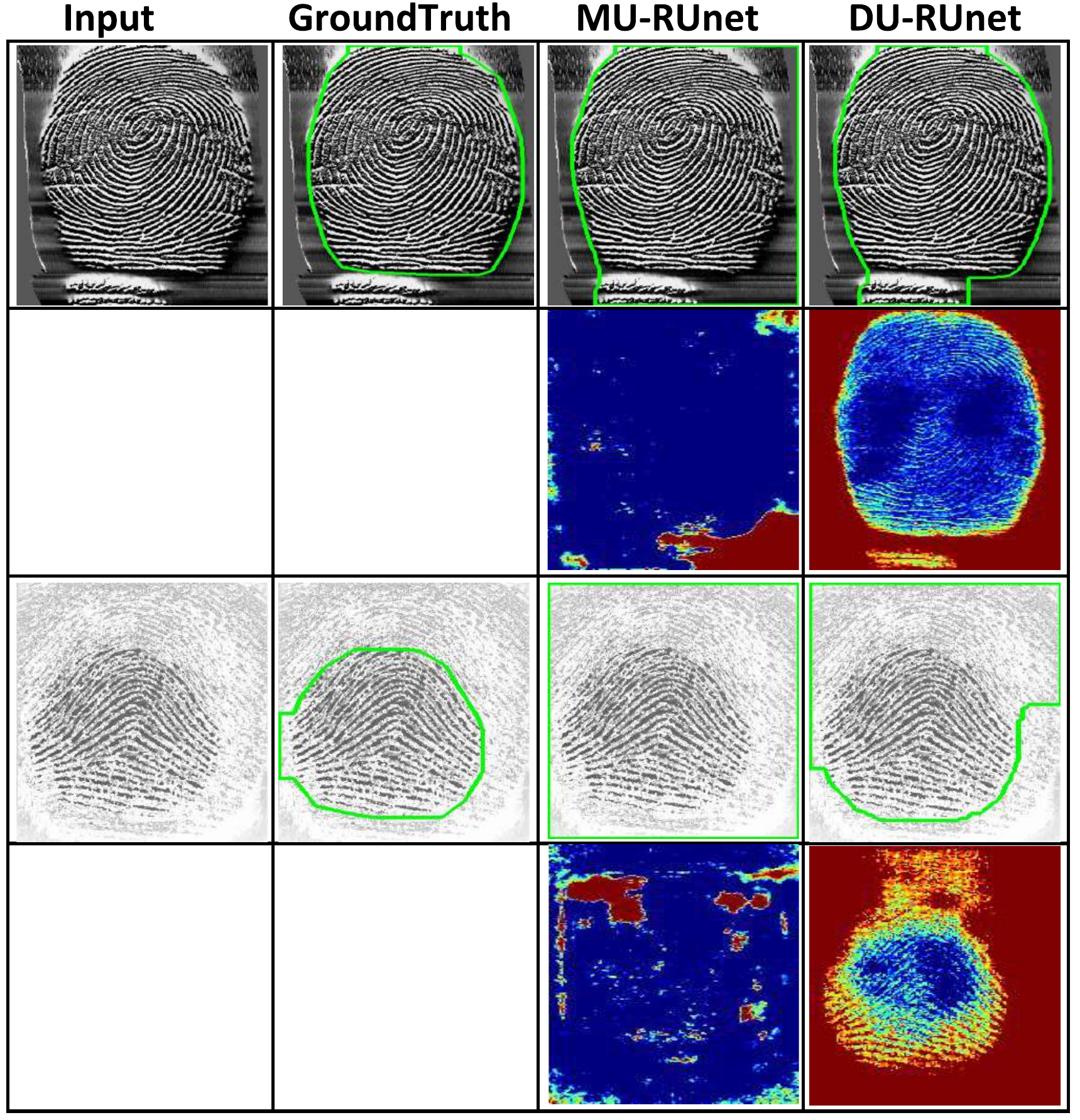}
\caption{\textbf{Visualization of model and data uncertainty} obtained while segmenting fingerprint roi. First and third rows depict the input fingerprint, segmented ground truth and the corresponding segmented images obtained using MU-RUnet~\cite{Joshi_2021_WACV} and proposed DU-RUnet. MU-RUnet is obtained after introducing Monte Carlo Dropout to capture model uncertainty, while \textbf{DU-RUnet} is designed to capture \textit{data uncertainty}. Predicted uncertainty is shown in the second and fourth row. The visualization of uncertainty values demonstrates the fact that the predicted model uncertainty only indicates high uncertainty under misclassified pixels, that too not well calibrated (blue and red color denote low and high uncertainty values respectively). On the other hand, predicted data uncertainty clearly discriminates noise and background pixels from the foreground which improves the robustness of the model towards noise.}
\label{visualization_introduction}
\end{figure}
\par 
Bayesian deep networks offer a practical way to identify noisy pixels in the input image through modelling the \textit{data uncertainty} arising due to sensor noise or occlusions. We demonstrate that modelling data uncertainty through a Bayesian framework helps the state-of-the-art fingerprint preprocessing models to learn noise-invariant features and improves their performance. Additionally, the predicted per-pixel data uncertainty serves as a tool for understanding of the human operators (see Figure \ref{visualization_introduction}).

\section{Related Work}

\subsection{Fingerprint ROI Segmentation} 
\subsubsection{Classical image processing based methods}Hu \textit{et al.}~\cite{seg3} and Thai \textit{et al.}~\cite{seg10} propose filtering based segmentation. Thai and Gottsclich~\cite{seg6} and Fahmy and Thabet~\cite{seg12} explore the potential of morphological operations for fingerprint roi segmentation. While Teixeira and Leite~\cite{seg8} and Raimundo \textit{et al.}~\cite{seg9} exploit the fingerprint ridge orientation information to segment foreground from background. 
\subsubsection{Learning based methods}Ferreira \textit{et al.}~\cite{seg4} and Yang \textit{et al.}~\cite{seg7} propose pixel-level clustering to discriminate foreground from background. While Liu \textit{et al.}~\cite{seg11}, Serafim \textit{et al.}~\cite{seg1} and Stojanovi{\'c} \textit{et al.}~\cite{seg2} propose patch level classification of foreground and background accompanied by postprocessing. 
\par None of the learning based architectures described above are end-to-end. Recently, Joshi \textit{et al.} \cite{Joshi_2021_WACV, joshi_sensor}, show that RUnet~\cite{wang2019recurrent} is an effective baseline for fingerprint roi segmentation. Furthermore, the authors incorporate Monte Carlo dropout to estimate model uncertainty and show that it helps to improve the performance of RUnet along with imparting model interpretability.
\subsection{Fingerprint Enhancement}
\subsubsection{Classical image processing based methods}Hong \textit{et al.} \cite{hong1998fingerprint}, Turroni \textit{et al.} \cite{turroni2012fingerprint}, Gottschlich and Sch{\"o}nlieb \cite{gottschlich2012oriented}, Gottschlich \cite{gottschlich2011curved} and Wang \textit{et al.} \cite{wang2008design} propose filtering in spatial domain.  Chikkerur \textit{et al.} \cite{chikkerur2007fingerprint} and Ghafoor \textit{et al.} \cite{ghafoor2014efficient} exploit information in Fourier domain. Sharma and Dey \cite{sharma2019two} propose a quality adaptive filtering in Fourier domain.
\subsubsection{Learning based methods}Schuch \textit{et al.} \cite{schuch2016convolutional} propose a deconvolutional auto-encoder (DeConvNet) to reconstruct poor quality fingerprints. Qian \textit{et al.} \cite{denseunet} propose DenseUnet while Wong and Lai~\cite{multitask_pr} and Li \textit{et al.} \cite{li2018deep} propose multi-tasking auto-encoder explicitly utilizing orientation field information. Joshi \textit{et al.}~\cite{indu2019wacv} propose a generative adversarial network (FP-E-GAN) for fingerprint enhancement. A detailed survey on fingerprint enhancement algorithms is conducted by Schuch \textit{et al.} \cite{schuch2017survey}.
\par Tiwari \textit{et al.} \cite{tiwari2014}, Vatsa \textit{et al.} \cite{vatsa2010} and Puri \textit{et al.} \cite{puri2010} evaluate the performance of state-of-the-art fingerprint matching system on the rural Indian population and conclude that it is challenging. Motivated by these works, we evaluate the enhancement performance of proposed work on challenging rural Indian fingerprints database.
\subsection{Uncertainty Estimation}
Predicting uncertainty makes learning based models trust worthy and useful from the perspective of safety~\cite{amodei2016concrete}. A wide range of approaches are proposed to estimate uncertainty using the Bayesian formulation  of neural networks such as Monte-Carlo Dropout~\cite{gal2016dropout}, Deep Ensembles~\cite{lakshminarayanan2017simple}, Maximum softmax probability~\cite{hendrycks2016baseline} and Stochastic Variational Bayesian Inference~\cite{louizos2017multiplicative}. These uncertainty prediction techniques are  successfully applied to detect out-of-distribution samples and misclassifications. Predictive uncertainty also finds its applications in active learning~\cite{kirsch2019batchbald}.

\par To summarize, uncertainty estimation serves as an effective tool that enables model understanding and robustness. It has been successfully utilized in various image processing applications~\cite{combalia2020uncertainty, kwon2020uncertainty,kurmi2019attending,Kurmi_2021_WACV_in}. The usefulness of estimating model uncertainty in fingerprint roi segmentation is recently explored \cite{Joshi_2021_WACV}. However, in principle, uncertainty can be either due to model weights (model uncertainty) or due to noise in the input (data uncertainty). Modelling data uncertainty is therefore especially useful to identify noisy regions and achieve robust performance on distorted and poor quality fingerprints.

\subsection{Research Contributions}
To the best of our knowledge, this research is the first work in fingerprints domain to predict data uncertainty and demonstrate its usefulness in preprocessing of fingerprints. To study its generalization ability, experiments are conducted on 13 publicly available fingerprint databases. The effect of modelling data uncertainty is studied on two tasks: fingerprint roi segmentation and enhancement, and three different network architectures. Furthermore, both qualitative and quantitative analysis of predicted data uncertainty is conducted to evaluate its effectiveness. Additionally, we also compare the model performance, inference time and predicted uncertainty after modelling data uncertainty versus the Monte Carlo dropout based model uncertainty. Visualizations of neural activations (using Seg-Grad-Cam~\cite{segcam}) and predicted uncertainty are illustrated to provide insights on the proposed work.
\begin{figure*}[!]
		\centering \includegraphics[width=\textwidth,height=5.5cm]{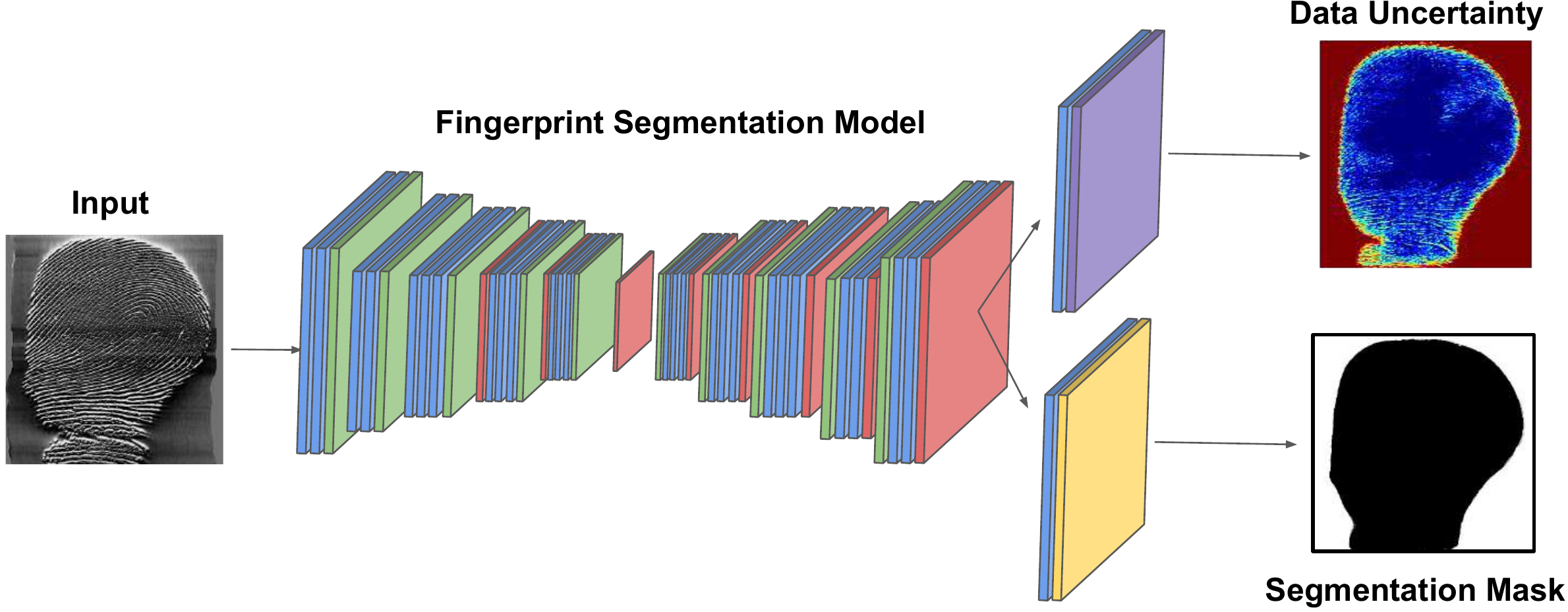}
		\caption{\textbf{Flowchart showcasing inference of data uncertainty.} The output layer comprises of two branches. For understanding, the case of fingerprint segmentation is shown where one branch predicts the segmentation mask whereas the other branch predicts the per-pixel data uncertainty.}
		\label{flowchart_data_uncertainty}
\end{figure*}
\section{Uncertainty in Fingerprint Preprocessing}
The success of deep models in fingerprint preprocessing mandates their use to obtain state-of-the-art performance. However, the standard fingerprint prepocessing models make predictions like a black-box and do not indicate when the model is highly likely to make an erroneous prediction. Uncertainty estimation provides a mechanism to understand what the model does not know and thus enables the end-users to separately handle more difficult cases or unreliable predictions. Baseline deterministic fingerprint preprocessing deep models are converted into Bayesian deep models to infer uncertainty from them. 
\par Uncertainty in a fingerprint preprocessing model can be primarily divided into two types: model uncertainty and data uncertainty.  On the other hand, data uncertainty captures the noise in the input fingerprint image due to factors such as dust and grease on the surface of fingerprint sensor, false traces arising during fingerprint acquisition, blurred ridges and unclear boundaries due to dry or wet fingertips. Data uncertainty cannot be reduced even if the model is trained on more training data. The usefulness of model uncertainty is recently studied by Joshi \textit{et al.} \cite{Joshi_2021_WACV}. In this research, we explain how to infer data uncertainty from a fingerprint preprocessing model and its benefits. 

\section{Estimating Data Uncertainty}
\label{data_uncertainty_estimation}
Data uncertainty is formalized as a probability distribution over model output. Given the input fingerprint image, data uncertainty estimation using Bayesian deep learning requires placing a prior distribution over output of model and calculating the variance of noise in model output. Predicted data uncertainty being input dependent, is learnt as a function of input image \cite{Gal_thesis}. To obtain both the preprocessed image and its associated uncertainty, network architecture of the baseline fingerprint preprocessing model is modified. Furthermore, since the background pixels are likely to be more noisy than foreground pixel, therefore per-pixel uncertainty is predicted.
\par As shown in Figure \ref{flowchart_data_uncertainty}, last layer of the baseline architecture is modified by splitting it into two. One branch predicts the model output (preprocessed image) whereas the other branch predicts the data uncertainty (noise variance). The mapping between input and preprocessed image is learnt in a supervised manner. However, no labels for uncertainty are used and the uncertainty values are learnt in an unsupervised manner. Furthermore, The loss function of the baseline architecture is also modified (as described in subsections \ref{regression} and \ref{classification}) to enable the modified architecture to learn to predict data uncertainty.

\par Fingerprint preprocessing models can be either based on regression or classification. In case of regression, the change in output can be directly calculated. However, in case of classification, in order to capture the true change in output, change in the values of logit is monitored rather than the change in output probabilities (output of softmax). Next, we describe the loss function to learn data uncertainty from both regression and classification based models.
\vspace{-0.1cm}
\subsection{Regression Based Models} \label{regression}
For a pixel $i$ of an input fingerprint image $x$, assuming the model output $f(x_{i})$ is corrupted with Gaussian zero mean random noise, estimating data uncertainty aims to learn the input dependent noise variance, $\sigma(x_{i})$. To learn data uncertainty, the original loss function $\frac{1}{n}\sum_{i=1}^{n}\|y_{i}-f(x_{i})\|^2$ is modified as follows:
	\begin{equation}
	\label{eq1}
	\begin{split}
	\frac{1}{n}\sum_{i=1}^{n} \frac{1}{2\sigma(x_{i})^2}\|y_{i}-f(x_{i})\|^2 + \frac{1}{2}\log\sigma(x_{i})^2
	\end{split}
\end{equation}
where $n$ denotes the total number of pixels in training images. Intuitively,  modifying the baseline architecture to predict data uncertainty and training with loss presented in equation \ref{eq1} enables it to adjust the residual error occuring on the noisy pixels by $\frac{1}{\sigma(x_i)}$ factor. Consequently, the model predicts high data uncertainty on noisy pixels. Furthermore, to ensure that the model does not predict high uncertainties for all pixels, the term $\log \sigma(x_i)^2$ is introduced. As a result, the modified loss function acts a \textit{noise-aware} loss.
\subsection{Classification Based Models} \label{classification}
To estimate data uncertainty from a classification model, the model is marginalized over the estimated data uncertainty in regression of logit space. For a pixel $i$ of an input fingerprint image $x$, let $f(x_{i})$ denotes the logit value before passing through softmax. Assuming $f(x_{i})$ is corrupted with Gaussian random noise with zero mean and variance $\sigma(x_{i})$, the network is optimized using Monte Carlo integration over cross-entropy loss for softmax probabilities of the sampled logits. As a result, the regular cross-entropy loss is modified as:
	\begin{equation}
	\label{eq2}
	\begin{split}
	\hat{x}_{i,t}=f(x_{i})+\sigma(x_{i})\hspace{1mm}\epsilon_{t}, \hspace{2mm}  \epsilon_{t}\sim \mathcal{N}(0,I)\\
		\frac{1}{n}\sum_{i=1}^{n}\log\frac{1}{T} \sum_{t=1}^{T} \exp(\hat{x}_{i,t,\hat{c}}-log \sum_{\hat{c}}\exp(\hat{x}_{i,t,\hat{c}}))
	\end{split}
\end{equation}
where $\hat{x}_{i,t}$ denotes the corrupted logit value for input $x_{i}$ at iteration $t$. $\hat{c}$, $n$ and $T$ denote the class label, total number of pixels in training images and number of Monte Carlo samples respectively. Similar to the case of regression, equation \ref{eq2} can be interpreted as learning a noise-aware loss.
\section{Databases}
\par To evaluate the effectiveness of of the proposed work, a wide range of challenging fingerprint databases in the public domain are used to conduct the experimental analysis. These databases are briefly described below:
\begin{enumerate}
    \item {Fingerprint Verification Challenge (FVC) Databases:} Three different FVC series 2000, 2002 and 2004 consisting of fingerprints acquired from different sensors, having varying background noise are used for this work. Each series has four databases and a well-defined training and testing set. Following the protocol, training and testing is conducted on a total of 960 and 9600 images respectively. The ground truth roi segmentation masks are obtained from \cite{seg6}\footnote{\protect {https://figshare.com/articles/dataset/Benchmark\_for\_Fingerprint\_Segment\\ation\_Performance\_Evaluation/1294209}}.
    \item {Rural Indian Fingerprint Database:} It has fingerprint samples collected from the rural Indian population extensively involved in manual work such as farmers, carpenters, villagers etc. It has 1631 fingerprint images acquired using an optical sensor.
\end{enumerate}

\section{Training and Testing}
Recurrent Unet (RUnet)~\cite{wang2019recurrent} is selected as the baseline architecture for fingerprint segmentation. It is a classification based model which is trained on cross-entropy loss. The architecture of RUnet is modified as suggested in Section \ref{data_uncertainty_estimation}. Modified architecture is named as \textit{DU-RUnet} (Recurrent Unet with Data Uncertainty). Training  and testing are performed on the respective training and testing subsets of FVC databases. The loss function presented in equation \ref{eq2} is used to train DU-RUnet. Hyper-parameter T=5 is used for training DU-RUnet. 
\par For fingerprint enhancement, two state-of-the-art fingerprint enhancement models: DeConvNet \cite{schuch2016convolutional} and FP-E-GAN~\cite{indu2019wacv} are modified to model data uncertainty. The resulting architectures are named \textit{DU-DeConvNet} and \textit{DU-GAN} respectively. Training is performed on synthetic dataset as suggested in~\cite{indu2019wacv} while testing us conducted on the Rural Indian Fingerprint Database \cite{puri2010}. Both of these baseline models have regression based loss function. Thus, for training DU-DeConvNet and DU-GAN, the training loss is modified as suggested in Section \ref{regression}. During testing of these modified architectures, to infer the preprocessed fingerprint image and the data uncertainty associated with it, only a single forward pass through the proposed architecture is required.

\section{Evaluation Metrics}
\subsection{Segmentation Performance}
\subsubsection{Dice and Jaccard Score} We employ two standard metrics: Dice \cite{dice1945measures} and Jaccard score \cite{choi2010survey} to assess the segmentation performance obtained by the proposed segmentation model compared to the ground truth roi segmentation masks. 
\par Although DU-RUnet is an end-to-end model, however, to have enough metrics for comparisons with state-of-the-art, we also evaluate DU-RUnet on impression 3 and 4 of FVC 2002-Db1a database over the patch based metrics described next.
\subsubsection{Erroneously Classified Patches}Let $patch_{1}$ represents a 16$\times$16 patch from predicted segmentation mask whereas $patch_{2}$ represents the corresponding ground truth patch manually marked by fingerprint experts. The percentage of erroneously classified patches (Err) is described as: 
\begin{equation}
	\begin{split}
	Err= \frac{\mbox{number of patches}(patch_{1}\neq patch_{2})}{\mbox{number of patches}(patch_{1})}
	\end{split}
\end{equation}

\subsubsection{Hit Coefficient and Mistake Coefficient} Hit coefficient (HC) and Mistake Coefficient (MC) indicate the relative foreground predicted correctly and incorrectly, respectively compared to the ground truth.
\begin{equation}
	\begin{split}
	HC= \frac{\mbox{Area}(P\cap G)}{\mbox{Area}(G)}\\
	MC= \frac{\mbox{Area}(P-G)}{\mbox{Area}(G)}
	\end{split}
\end{equation}
where $P$ and $G$ represent the foreground fingerprint area in the predicted segmentation mask and ground truth segmentation.
\subsection{Enhancement Performance}
\subsubsection{Fingerprint Quality Assessment}
To quantify the improvement in fingerprint quality after enhancement, we calculate fingerprint image quality scores using \textit{Nfiq} module of NBIS \cite{nbis}. Nfiq returns a score in the range [1,5] where 1 and 5 signify the best and the worst fingerprint quality.
\subsubsection{Ridge Reconstruction Ability} In order to evaluate the ridge reconstruction ability of the proposed DU-GAN, we calculate Peak signal-to-noise ratio (PSNR) between the enhanced image generated by DU-GAN and the ground truth binarized fingerprint image obtained using binarization module of NBIS. However, since the ground truth binarization cannot be reliably generated on the testing database, synthetic distorted fingerprint images are generated for this experiment. Good quality synthetic fingerprints are generated using \cite{anguli} at first, which are then degraded using various noise and background variations.
\subsubsection{Matching Performance} To demonstrate the improved fingerprint matching performance, we report the average Equal Error Rate (EER) and plot the Detection Error Tradeoff (DET) curve. Fingerprint matching systems used are Bozorth \cite{nbis} and MCC \cite{mcc10}, \cite{mcc11}, \cite{mcc12},  \cite{ferrara2014two}.
\section{Results and Discussions}
\subsection{Data Uncertainty Guides Noise-aware Segmentation}
\begin{figure*}[!]
		\centering \includegraphics[width=0.9\textwidth,height=6.2cm]{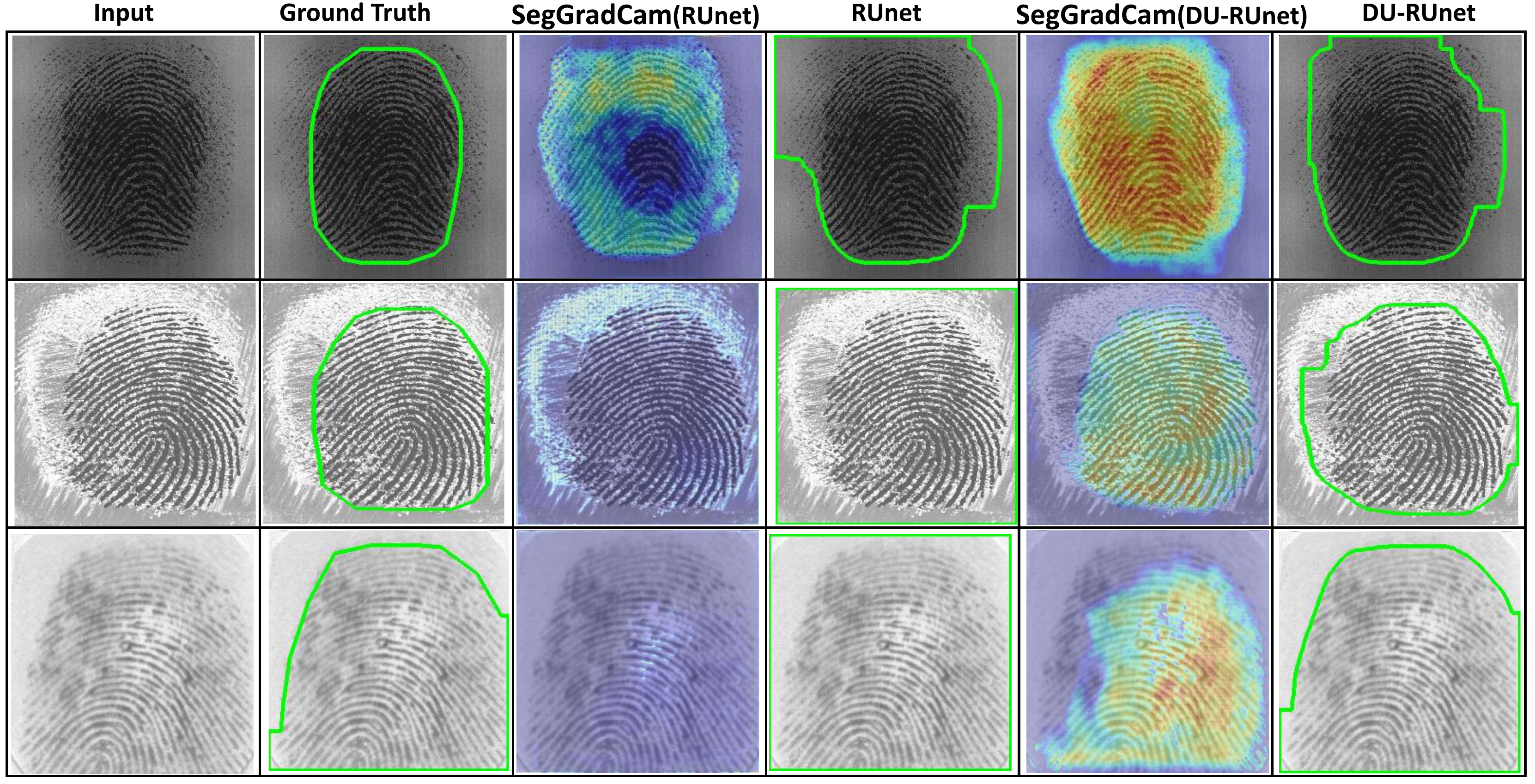}
		\caption{\textbf{Visualizations obtained using Seg-Grad-Cam} (best viewed in colour). Higher activations around the foreground and boundaries are obtained by DU-RUnet compared to the baseline RUnet. This explains the improved segmentation performance by RUnet on noisy background pixels after modelling data uncertainty.}
		\label{segcam_data_uncertainty}
\end{figure*}

Table \ref{table_runet_vs_du_runet} reports the improved dice and jaccard scores obtained by the proposed DU-RUnet as compared to baseline RUnet. To fathom reasons for the same, Figure \ref{segcam_data_uncertainty} showcases sample visualizations obtained for RUnet and DU-RUnet using Seg-Grad-Cam \cite{segcam}. Results reveal that predicting data uncertainty helps the model to identify noisy regions in fingerprint images due to which higher activations are obtained around foreground fingerprint pixels. As a result, improved segmentation performance on noisy background pixels is obtained.
\begin{table}[]
\caption{Comparison Of Jaccard Similarity And Dice Score Obtained By Baseline RUnet And Proposed DU-RUnet.}
\centering
\begin{tabular}{|c||c|c||c|c|}
\hline
\multirow{2}{*}{\textbf{Database}} & \multicolumn{2}{l|}{\textbf{Jaccard Similarity ($\uparrow$) }} & \multicolumn{2}{l|}{\textbf{Dice Score ($\uparrow$) }} \\ \cline{2-5} 
               &     \textbf{RUnet}     &     \textbf{DU-RUnet}       & \textbf{RUnet}          &       \textbf{DU-RUnet}     \\ \hline
2000DB1  &88.15& \textbf{88.52} &93.34& \textbf{93.62}  \\
2000DB2 &86.40& \textbf{88.07}  &92.39& \textbf{93.42} \\
2000DB3  &93.74& \textbf{95.36} &96.50& \textbf{97.55} \\
2000DB4 &94.28& \textbf{94.97}  &97.04& \textbf{97.40} \\
2002DB1   &96.95& \textbf{97.07} &98.44 & \textbf{98.50} \\
2002DB2   &94.88& \textbf{95.43} &97.28& \textbf{97.60}  \\
2002DB3  &91.83&\textbf{93.06}  &95.53& \textbf{96.25} \\
2002DB4   &91.17& \textbf{91.89} &95.32& \textbf{95.74}  \\
2004DB1   &98.78& \textbf{99.00} &99.38& \textbf{99.50}  \\
2004DB2   &93.94& \textbf{96.37}  &96.69& \textbf{98.14} \\
2004DB3   &94.62&\textbf{95.47} &97.17& \textbf{97.65} \\
2004DB4  &94.73 & \textbf{95.61} &97.21 &\textbf{97.70} \\    \hline
\end{tabular}
\label{table_runet_vs_du_runet}
\end{table}

\subsection{Comparison of Model and Data Uncertainty}
To provide insights on what exactly data uncertainty captures and how it is different than model uncertainty \cite{Joshi_2021_WACV}, we perform a detailed comparison of the proposed DU-RUnet (RUnet with data uncertainty) with the recently proposed MU-RUnet \cite{Joshi_2021_WACV} (RUnet with model uncertainty).
\par Table \ref{table_mu_runet_vs_du_runet} and Table \ref{table_sota_segmentation} compare the segmentation performance obtained by MU-RUnet and DU-RUnet. DU-RUnet outperforms MU-RUnet on majority of the databases. These results demonstrate the fact that data uncertainty turns to be more useful than model uncertainty in facilitating correct segmentation of noisy background pixels.
\par
\begin{figure*}[!]
		\centering \includegraphics[width=0.9\textwidth,height=6.2cm]{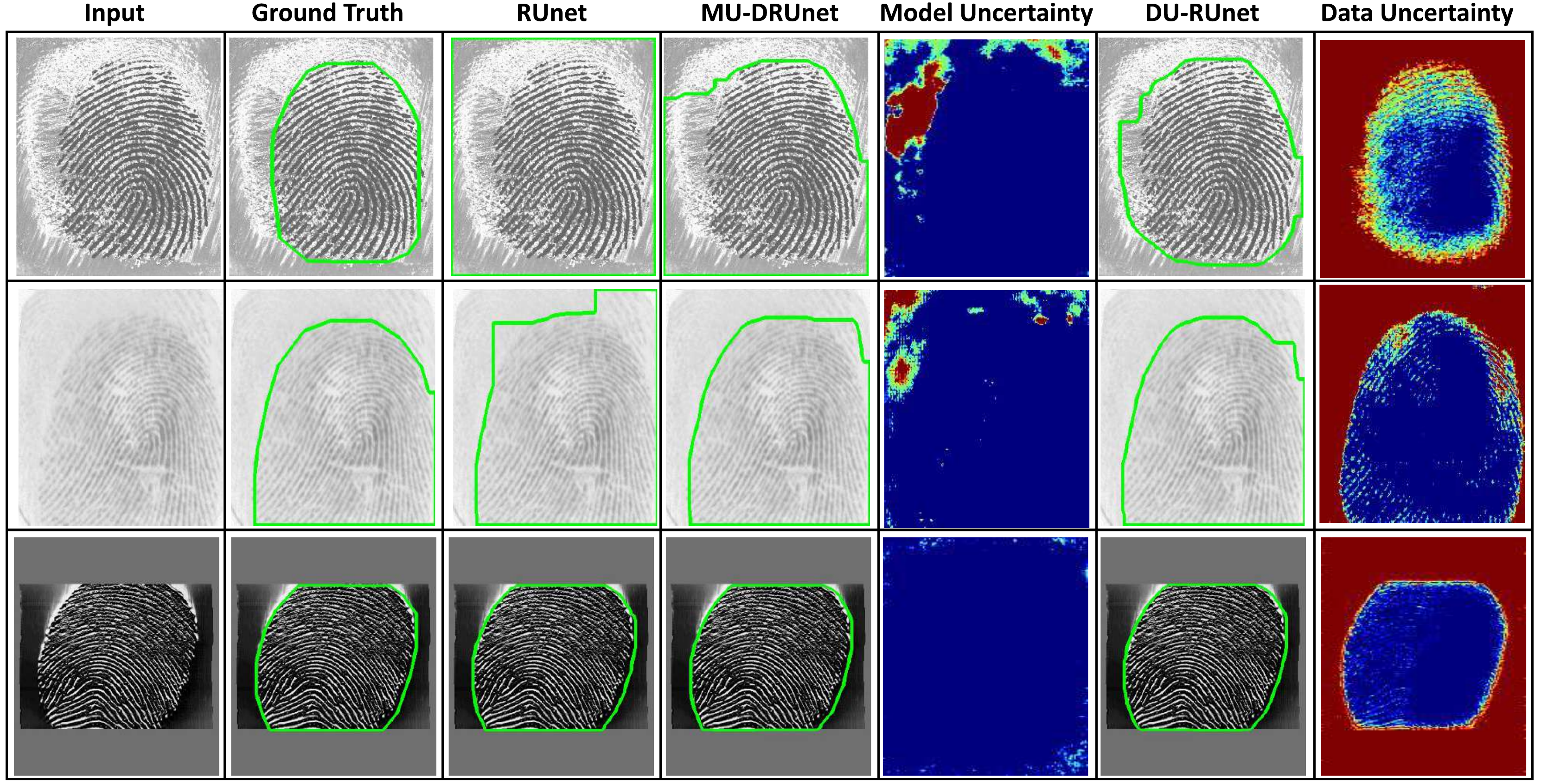}
		\caption{\textbf{Visualization of model and data uncertainty}. Sample cases demonstrating the fact that predicting either of the two kind of uncertainties improves the segmentation performance as both of these capture different but useful information. Model uncertainty captures model's confidence in prediction. As a result, higher uncertainty around incorrect predictions is obtained compared to the correctly predicted pixels. Data uncertainty on the other hand, captures the noise in in the fingerprint image. Consequently, higher data uncertainty is predicted around background and boundaries as compared to the foreground.}
		\label{model_versus_data_uncertainty}
		\vspace{-8pt}
\end{figure*}

 Next, to analyze the interpretability of predicted uncertainties, we plot Figure \ref{model_versus_data_uncertainty}. Sample cases demonstrate the fact that indeed predicting either type of uncertainty helps to improve the baseline segmentation performance. Both of these uncertainties capture complementary information. Model uncertainty captures model's confidence in prediction due to which higher uncertainty values are obtained for incorrectly classified pixels. On the other hand, data uncertainty captures noise in the input fingerprint image due to which higher uncertainties values are obtained around noisy and background pixels as compared to the foreground. Furthermore, consistent with the literature \cite{Gal_thesis}, we observe data uncertainty values to be better calibrated than model uncertainties.

\begin{table}[]
\caption{Comparison Of Jaccard Similarity And Dice Score Obtained After Incorporating Model And Data Uncertainty.}
\centering
\begin{tabular}{|c||c|c||c|c|}
\hline
\multirow{2}{*}{\textbf{Database}} & \multicolumn{2}{l|}{\textbf{Jaccard Similarity ($\uparrow$) }} & \multicolumn{2}{l|}{\textbf{Dice Score ($\uparrow$) }} \\ \cline{2-5} 
               &     \textbf{MU-RUnet}     &     \textbf{DU-RUnet}       & \textbf{MU-RUnet}          &       \textbf{DU-RUnet}     \\ \hline
2000DB1  &87.97& \textbf{88.52} &93.14& \textbf{93.62}  \\
2000DB2 &\textbf{88.43}& 88.07  &\textbf{93.58}& 93.42 \\
2000DB3  &\textbf{95.39}& 95.36 &\textbf{97.57}& 97.55 \\
2000DB4 &94.89& \textbf{94.97}  &97.36& \textbf{97.40} \\
2002DB1   &96.83& \textbf{97.07} &98.38 & \textbf{98.50} \\
2002DB2   &95.13& \textbf{95.43} &97.40& \textbf{97.60}  \\
2002DB3  &\textbf{93.87}&93.06  &\textbf{96.73}& 96.25 \\
2002DB4   &91.53& \textbf{91.89} &95.54& \textbf{95.74}  \\
2004DB1   &98.88& \textbf{99.00} &99.49& \textbf{99.50}  \\
2004DB2   &95.98& \textbf{96.37}  &97.93& \textbf{98.14} \\
2004DB3   &95.29&\textbf{95.47} &97.55& \textbf{97.65} \\
2004DB4  &\textbf{96.18} & 95.61 &\textbf{98.03} & 97.70 \\  
\hline
\end{tabular}
\label{table_mu_runet_vs_du_runet}
\end{table}
\begin{table}
\caption{ Comparison of Segmentation Performance Obtained By DU-RUnet and MU-RUnet.}
	\centering	
\begin{tabular}{|p{2.8cm}|p{1cm}|p{1cm}|p{1cm}|}
		\hline
		\textbf{Algorithm}&\textbf{Err ($\downarrow$)}&\textbf{HC($\uparrow$) }&\textbf{MC ($\downarrow$)}\\
		\hline\hline
MU-RUnet \cite{Joshi_2021_WACV}  & 0.0173& \textbf{0.9949}&0.0313\\
\textbf{DU-RUnet (Proposed)}  & \textbf{0.0163}& 0.9936&\textbf{0.0301}\\
\hline
	\end{tabular}
	\label{table_sota_segmentation}
\end{table}
\begin{table}
\caption{Comparison of inference time.}
    \centering    
    \begin{tabular}{|c|c|}
        \hline
        \textbf{Architecture}&\textbf{Time (sec.)}\\
        \hline\hline
        RUnet \cite{wang2019recurrent}&0.22\\
        MU-RUnet \cite{Joshi_2021_WACV}&1.03\\
        \textbf{DU-RUnet (Proposed)} &0.22\\
        \hline
    \end{tabular}
    \label{table_time_drunet}
\end{table}

\par Lastly, Table \ref{table_time_drunet} compares the inference time for MU-RUnet and DU-RUnet. System configuration on which inference time is computed consists of a Tesla V100 GPU and a Xeon Silver 4215 CPU. Please note that the inference time of MU-RUnet depends upon the number of samples used for the Monte Carlo integration. In this study, MU-RUnet uses five samples during testing. As reported in Table \ref{table_time_drunet}, inference time of DU-RUnet is comparable to RUnet. However, due to Monte Carlo Integration, inference time of MU-RUnet is approximately five times of DU-RUnet.

\subsection{Analysis of Data Uncertainty}
\subsubsection{Qualitative Analysis}
Figure \ref{model_versus_data_uncertainty} presents sample input images and the data uncertainty predicted by the proposed DU-RUnet. High uncertainty is predicted for background as compared to the foreground. Furthermore, boundary pixels around the input fingerprint image in the top row are far more noisy as compared to second and third row. Consequently, the predicted uncertainty around the boundaries is higher for the top row compared to the second and third row. These results demonstrate the reliability of the predicted data uncertainty.

\subsubsection{Quantitative Analysis}
\begin{figure}
\centering
\subfigure [] {\includegraphics[height=3.25cm]{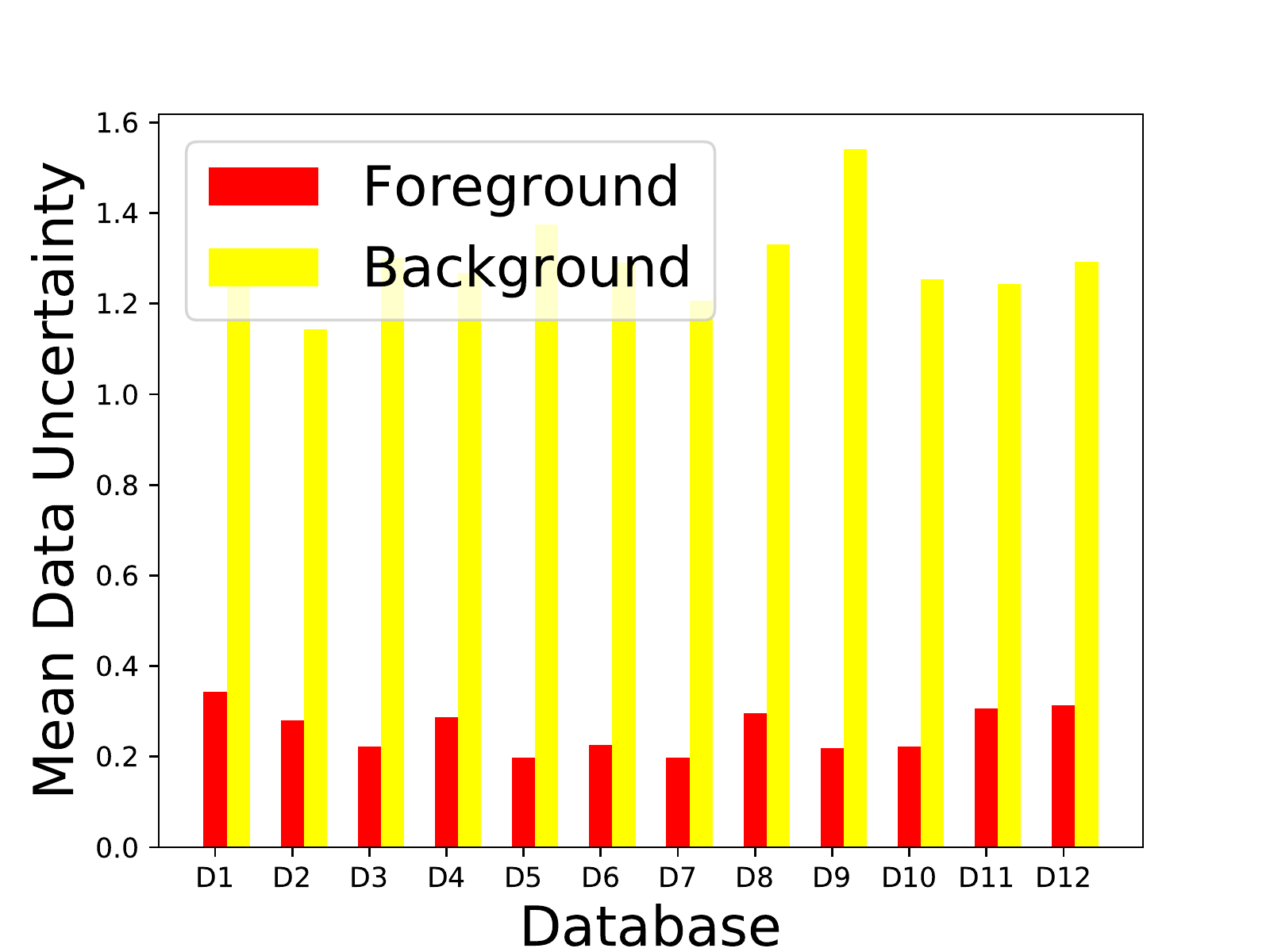}}
\subfigure [] {\includegraphics[height=3.25cm]{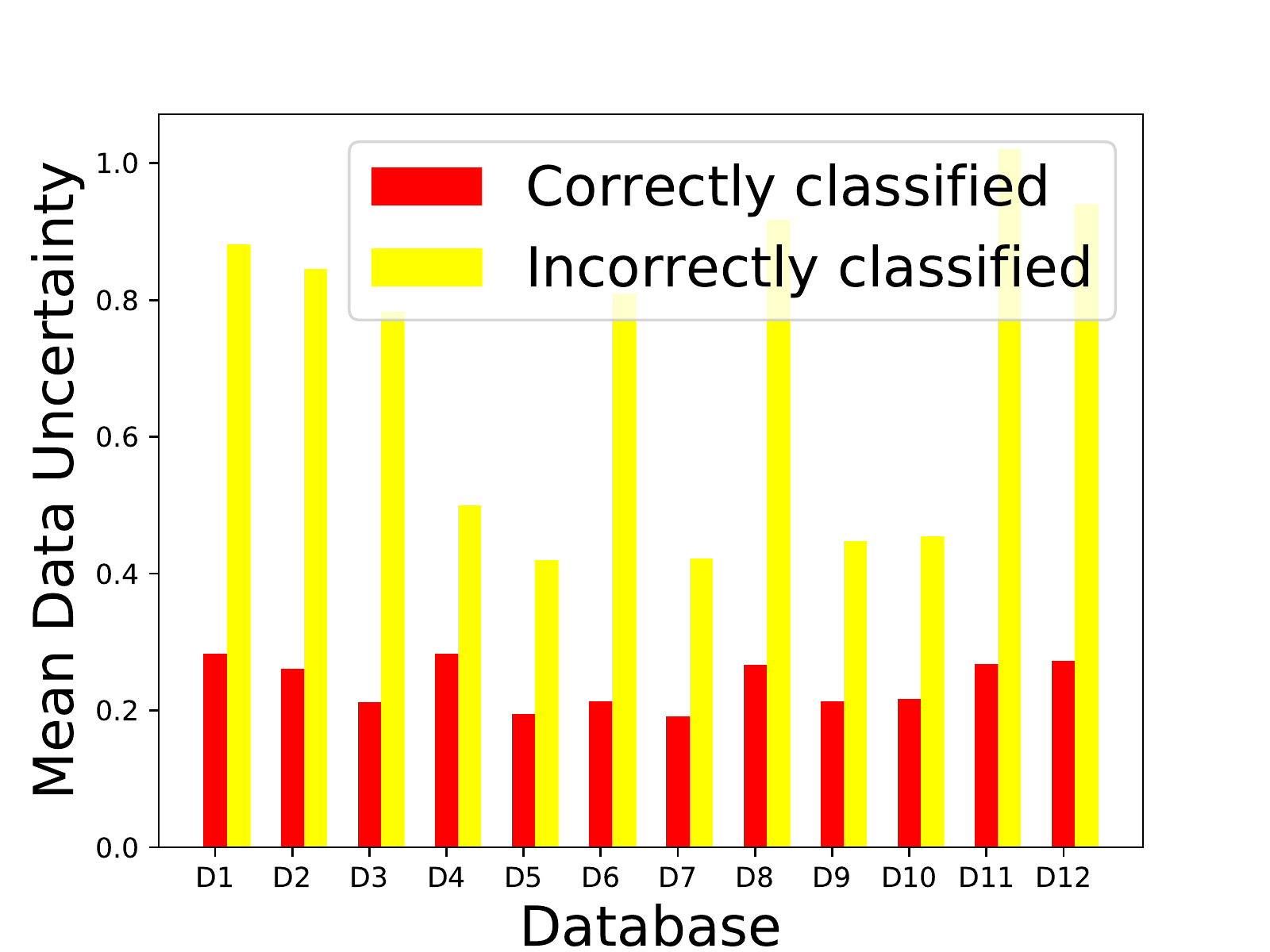}}
    \caption{\textbf{Comparison of predicted data uncertainty} for (a) foreground and background pixels (b) correctly and incorrectly classified pixels. Higher mean uncertainty obtained for background and incorrectly classified pixels demonstrates the efficacy of data uncertainty prediction. D1 to D12 represent FVC2000 DB1 to FVC2004 DB4 respectively (in order).}
\label{data_uncertainty}
\end{figure}

For quantitatively demonstrating the efficacy of data uncertainty predicted by the proposed DU-RUnet, Figure \ref{data_uncertainty} illustrates the mean data uncertainty predicted for: background (with respect to ground truth roi mask) versus foreground pixels and correctly versus incorrectly classified pixels. As expected, the mean uncertainty predicted for background is significantly higher compared to foreground. Likewise, the mean uncertainty predicted for incorrectly classified pixels is way higher compared to correctly classified pixels. These results verify the claim that DU-RUnet predicts high data uncertainty around boundaries and noisy background pixels. 

\subsection{Generalization Ability}
All the experimental analysis presented so far is conducted on fingerprint ROI segmentation. To establish the effectiveness of modelling data uncertainty in fingerprint preprocessing, in general, we also demonstrate its impact in fingerprint enhancement. In this direction, we take two state-of-the-art fingerprint enhancement models DeConvNet \cite{schuch2016convolutional} and FP-E-GAN~\cite{indu2019wacv} and modify them to DU-DeConvNet and DU-GAN to obtain data uncertainty from these baseline architectures.

\par
\begin{figure}
\centering
\subfigure [] {\includegraphics[height=3.25cm]{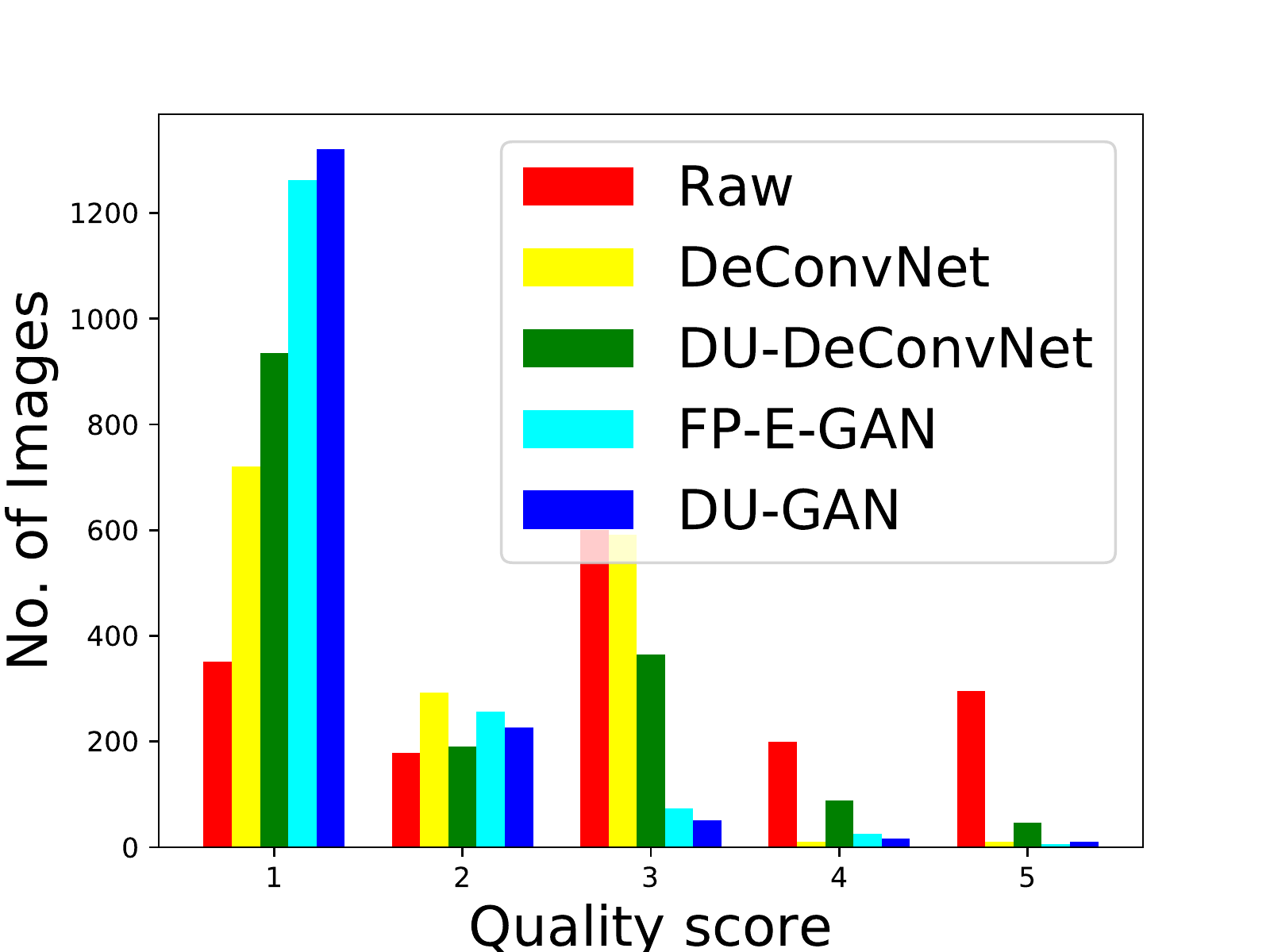}}
\subfigure [] {\includegraphics[height=3.25cm]{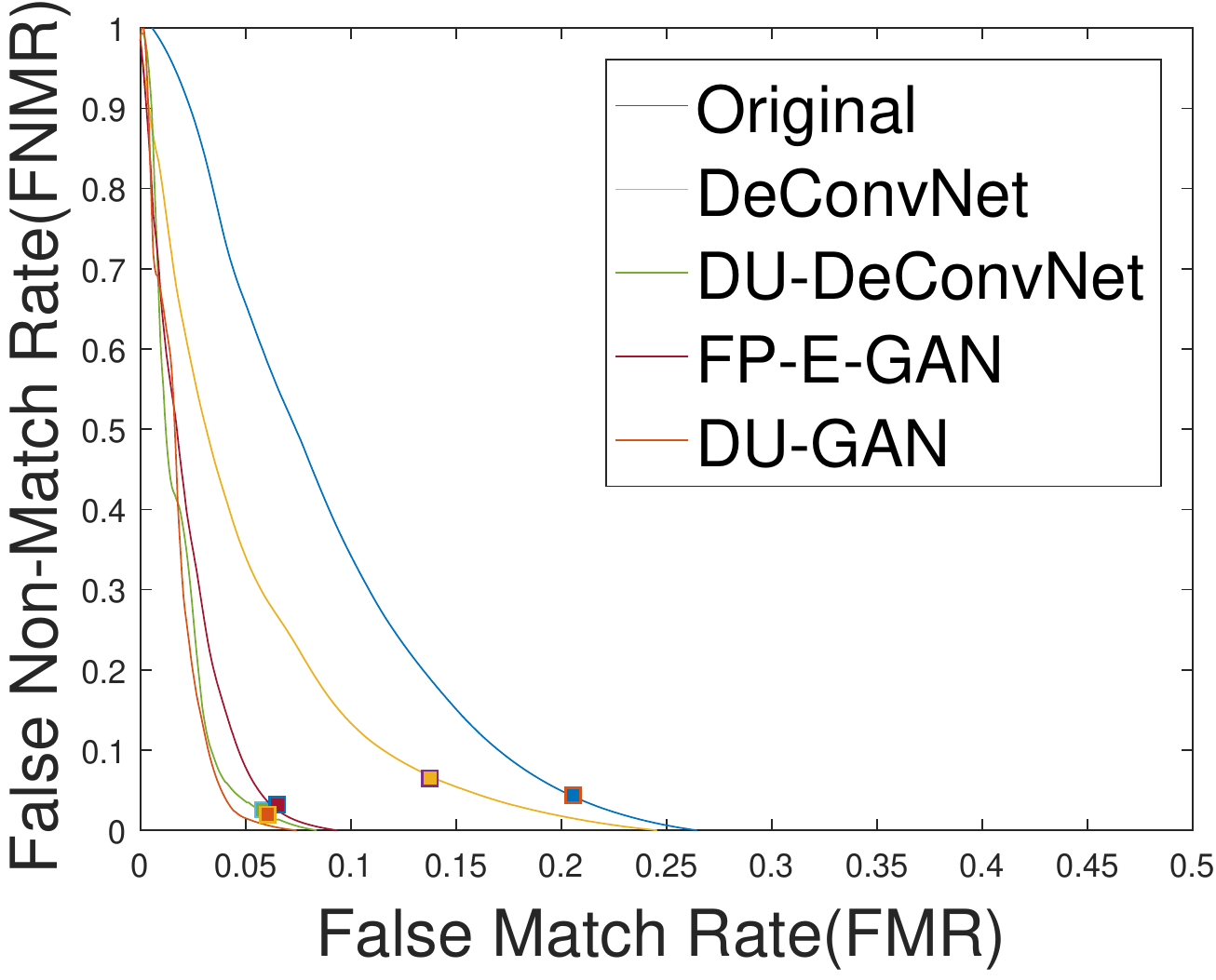}}
    \caption{\textbf{Improved enhancement performance} obtained on by the proposed DU-DeConvNet and DU-GAN (after modelling data uncertainty) demonstrated through (a) Improved Nfiq quality scores (lower is better) (b) DET curve demonstrating reduced EER while performing matching using MCC matcher.  }
\label{iiitd_rural}
\end{figure}

 Table \ref{table_nfiq_biometrics} and Figure \ref{iiitd_rural} (a) demonstrate the fact that the fingerprint quality scores are improved for both the baseline enhancement models after modifying them to predict data uncertainty. Likewise, as indicated in Table \ref{table_eer_biometrics_iiitd} and Figure \ref{iiitd_rural} (b), images generated by DU-DeConvNet and DU-GAN obtain better matching performance compared to the baseline DeConvNet and FP-E-GAN.

\par
\begin{figure}
\centering \includegraphics[width=0.48\textwidth,height=4.5cm]{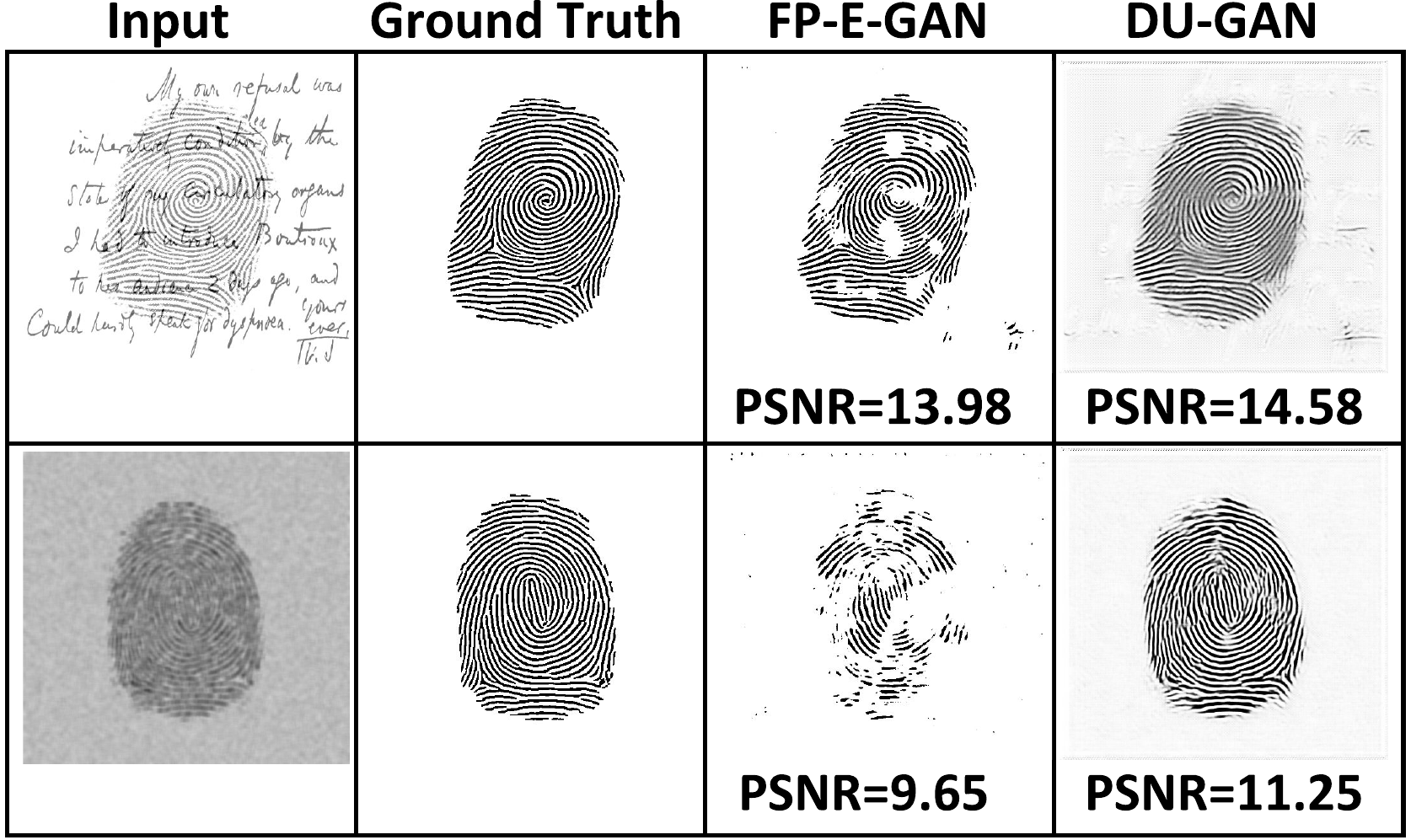}
\caption{Sample test cases showcasing the \textbf{improvement in ridge reconstruction ability} of FP-E-GAN after modelling data uncertainty, resulting in proposed DU-GAN.}
\label{data_uncertainty_psnr}
\end{figure}

 Next, we show that the ridge reconstruction ability is indeed improved after modelling data uncertainty. Figure \ref{data_uncertainty_psnr} compares the PSNR value obtained by DU-GAN (as it is better performing architecture than DU-DeConvNet) and its corresponding baseline architecture, FP-E-GAN. Higher PSNR value with respect to the ground truth binarized image is obtained for DU-GAN which signifies that DU-GAN performs better than baseline FP-E-GAN in reconstructing the distorted ridges.

\par
\begin{figure}
		\centering \includegraphics[width=0.48\textwidth,height=5cm]{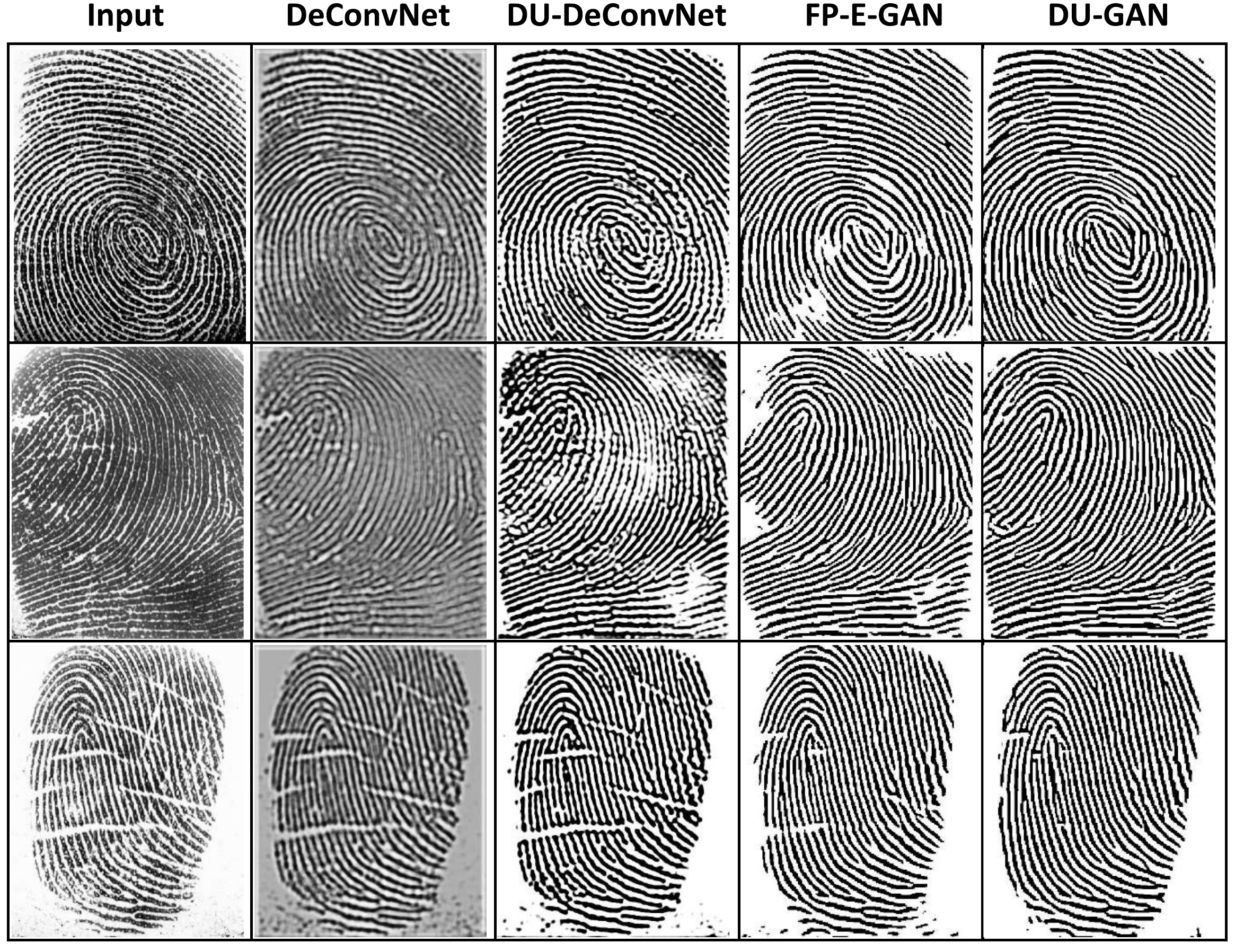}
		\caption{Sample challenging cases showcasing improved performance by state-of-the-art fingerprint enhancement algorithms after modelling data uncertainty.}
		\label{data_uncertainty_gan}
		\vspace{-8pt}
\end{figure}

 Lastly, Figure \ref{data_uncertainty_gan} showcases the improvement in enhancement performance after modifying the state-of-the-art fingerprint enhancement models to predict data uncertainty. We observe that DU-ConvNet and DU-GAN perform far better than corresponding baselines in predicting missing ridge information and improving the overall ridge-valley clarity.

\begin{table}
\caption{Average Nfiq Quality Scores Obtained On Rural Indian Fingerprint Database.}
    \centering    
    \begin{tabular}{|c|c|}
        \hline
        \textbf{Enhancement Algorithm}&\textbf{Avg. Nfiq Score ($\downarrow$)}\\
        \hline\hline
        Raw Image & 2.94\\
        DeconvNet \cite{schuch2016convolutional}& 1.95\\
        \textit{DU-DeConvNet} &\textit{1.84}\\
        FP-E-GAN \cite{indu2019wacv}&1.31\\
        \textbf{DU-GAN} &\textbf{1.26}\\
        \hline
    \end{tabular}
    \label{table_nfiq_biometrics}
\end{table}

\begin{table}
\caption{Average EER Obtained On Rural Indian Fingerprint Database.}
	\centering	
	\begin{tabular}{|p{2cm}|p{1.4cm}|p{1.6cm}|}
		\hline
		\textbf{Enhancement Algorithm}&\textbf{Matching Algorithm}&\textbf{Avg. EER ($\downarrow$)}\\
		\hline\hline
		Raw Image & Bozorth&16.36\\
		DeConvNet \cite{schuch2016convolutional} & Bozorth& 10.93\\
		\textit{DU-DeConvNet} & Bozorth&\textit{8.71}\\
		FP-E-GAN \cite{indu2019wacv} & Bozorth& 7.30\\
		\textbf{DU-GAN} & Bozorth&\textbf{7.13}\\
		\hline\hline
		Raw Image & MCC& 13.23\\
		DeConvNet \cite{schuch2016convolutional} & MCC& 10.86\\
		FP-E-GAN \cite{indu2019wacv} & MCC& 5.96\\
		\textit{DU-DeConvNet}& MCC&\textit{5.36}\\
		\textbf{DU-GAN}& MCC&\textbf{5.13}\\
		\hline
	\end{tabular}
	\label{table_eer_biometrics_iiitd}
\end{table}

\section{Conclusion and Future Work}
This research is the first work in the fingerprints domain to demonstrate the effectiveness of modelling data uncertainty through a deep Bayesian network. Proposed methodology is tested on fingerprint roi segmentation and enhancement. Extensive experimentation over a wide range of databases and network architectures showcases the generalization ability of the proposed work. Insights on the improved model performance are provided through visualization of neural activations. Furthermore, qualitative and quantitative analysis of predicted data uncertainty is conducted which confirms that the higher data uncertainty is predicted around noisy and background pixels compared to clear foreground region. A detailed comparison between model performance obtained after incorporating model uncertainty and data uncertainty is conducted. Results reveal that modelling both the type of uncertainty is helpful as both the uncertainties capture different but useful information. However, the time taken to infer data uncertainty is much lower compared to the time required to infer model uncertainty. In future, the usefulness of uncertainty information in other stages of fingerprint matching pipeline can be studied.
\section*{Acknowledgment}
Authors thank the HPC facility of Inria Sophia Antipolis and IIT Delhi for computational resources used in this research. This work is partly supported by the French Government (National Research Agency, ANR) under grant agreement ANR-18-CE92-0024. I. Joshi is partially supported by the Raman-Charpak Fellowship 2019.

\bibliographystyle{ieee}
\bibliography{reference}

\end{document}